\documentclass[10pt,twocolumn,letterpaper]{article}

\usepackage{cvpr}
\usepackage{times}
\usepackage{epsfig}
\usepackage{graphicx}
\usepackage{amsmath}
\usepackage{amssymb}
\usepackage{booktabs}
\usepackage{array}
\usepackage{multirow}
\usepackage[table,x11names]{xcolor}
\usepackage{soul}
\usepackage{bm}

\definecolor{LightCyan}{rgb}{0.88,1,1}

\usepackage[pagebackref=true,breaklinks=true,letterpaper=true,colorlinks,bookmarks=false]{hyperref}

\cvprfinalcopy 

\def\cvprPaperID{1050} 

\ifcvprfinal\fi
\begin{document}

\title{Semantic Regularisation for Recurrent Image Annotation}

\author{Feng Liu$^{1,2}$ \quad Tao Xiang$^2$ \quad Timothy M. Hospedales$^3$ \quad Wankou Yang$^1$ \quad Changyin Sun$^1$\\
$^1$Southeast University, China 
\\$^2$Queen Mary, University of London, UK \quad  $^3$University of Edinburgh, UK \\
{\tt\small \{liufeng,wkyang,cysun\}@seu.edu.cn,\quad \{feng.liu,t.xiang\}@qmul.ac.uk,\quad t.hospedales@ed.ac.uk}
}

\maketitle

\begin{abstract}

The ``CNN-RNN'' design pattern is increasingly widely applied in a variety of image annotation tasks including  multi-label classification and captioning. Existing models use the weakly semantic CNN hidden layer or its transform as the image embedding that provides the interface between the CNN and RNN. This leaves the RNN overstretched with two jobs: predicting the visual concepts and modelling their correlations for generating structured annotation output. Importantly this makes the end-to-end training of the CNN and RNN slow and ineffective due to the difficulty of back propagating gradients through the RNN to train the CNN. We propose a simple modification to the design pattern that makes learning  more effective and efficient. Specifically, we propose to use a semantically regularised embedding layer as the interface between the CNN and RNN. Regularising the interface can partially or completely decouple the learning problems, allowing each to be  more effectively trained and jointly training much more efficient. Extensive experiments show that  state-of-the art performance is achieved on multi-label classification  as well as image captioning. 
\end{abstract} 

\vspace{-0.5cm}
\section{Introduction}

The classic task of image recognition is beginning to approach a solved problem with the latest Inception-ResNet \cite{inceptionv4} achieving a top 5 error rate of $3.08\%$ on the ILSVRC15 \cite{ilsvrc} dataset, surpassing humans. Interest is therefore growing in generating richer descriptions of image properties rather than simple categorisations, including multi-label classification/tagging \cite{sinn,feifei,basic_model,cnnrnn} and image captioning \cite{nic,mil,feifeicap,Xu2015show,attention,Wu_2016_CVPR}. 

In multi-label classification  the aim is to describe rather than merely recognise an image by annotating all visual concepts that appear in the image. The label space is thus richer than in the single-label  recognition case -- labels can refer to scene properties, objects, attributes, actions, aesthetics \etc. Such labels have richer relationships, e.g., a policeman is a person; car and sky co-exist more often than car and sea. Image captioning has a related aim, with the difference of producing a complete natural language sentence description conditioned on the image content, rather than a simple unordered set of labels. For both problems an effective model needs to fulfil two closely-related tasks well: predicting a set of visual concept labels and modelling  inter-label correlations.  For  label-correlation modelling,  structured learning strategies are typically employed, which in the case of multi-label classification helps  to better distinguish visually ambiguous concepts as well as suppress false predictions (e.g., modelling the car-sky-sea correlation can rectify false prediction of sea in place of sky when a car is present).  For image captioning, structured learning is even more critical to generate an ordered list of words that encode a valid as well as relevant sentence. 


\begin{figure*}[htbp]
\begin{center}
\includegraphics[width=140mm]{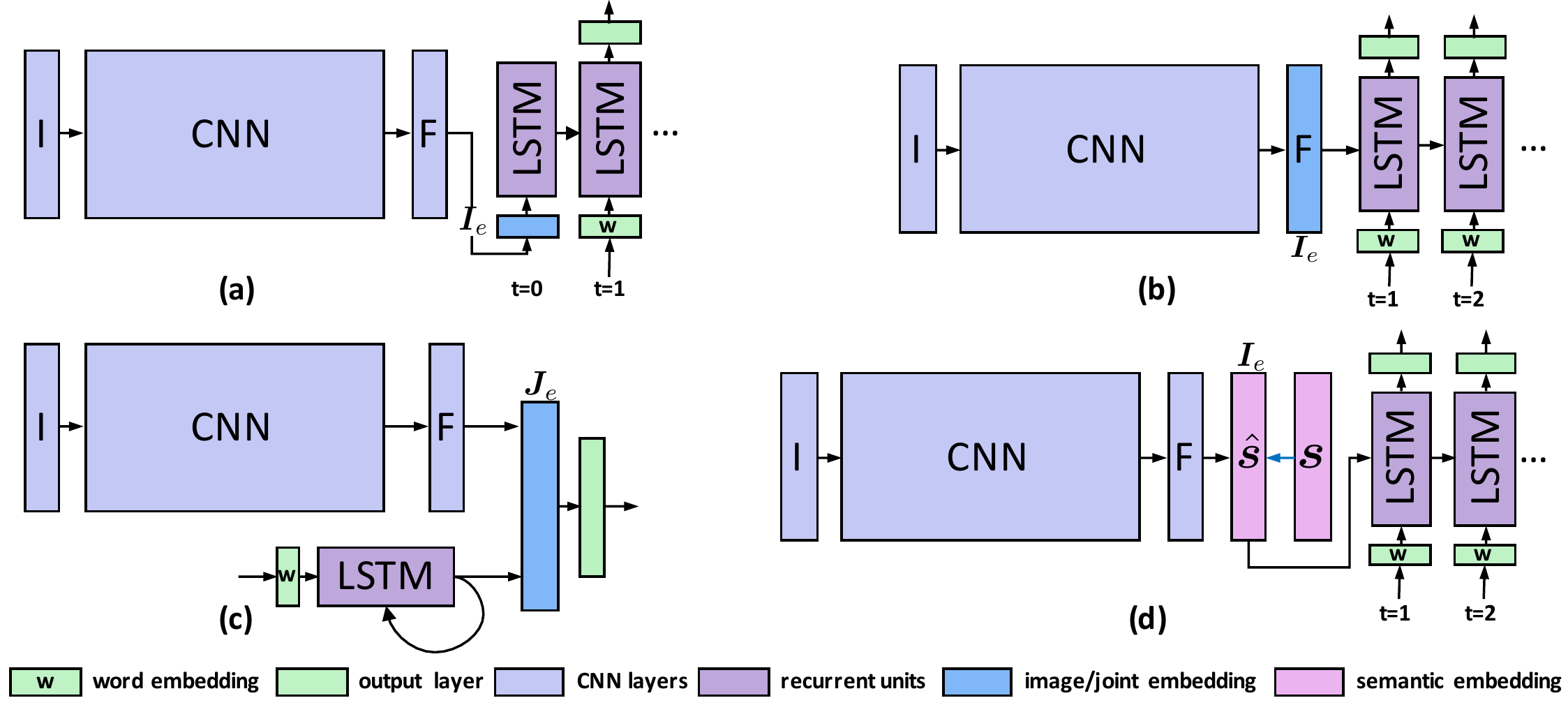} 
\caption{CNN-RNN architectures for image annotation (multi-label classification and captioning). In all models, LSTM is used as the RNN model. (a)  CNN encodes an image (I) to a feature representation (F). The image embedding $I_{e}$ and word representation go through the same word embedding layer before being fed into the LSTM \cite{showandtell}. (b) The image CNN output features F  set the LSTM hidden states  \cite{basic_model}. (c) The image CNN output feature layer is integrated with the LSTM output via late fusion \cite{mrnn,cnnrnn}. (d) The proposed semantically regularised model. The CNN model is regularised by the ground truth semantic concepts $\bm{s}$, which serve as strong deep supervision to guide the learning of the CNN layers. The CNN prediction layer $\hat{\bm{s}}$ is used as image embedding which is used to set the LSTM  initial states. Best viewed in colour.} \label{fig:compare}
\end{center}
\vspace{-5mm}
\end{figure*}

Recently, the convolutional neural network -- recurrent neural network (CNN-RNN)  encoder-decoder design pattern has become popular to address the structured label prediction task in both multi-label classification \cite{basic_model,cnnrnn} and image captioning \cite{showandtell,nic,Xu2015show,attention}. A CNN is used to encode the image into a fixed length vector, which is then fed into an RNN that either decodes it into a list of tags (multi-label) or sequence of words composing a sentence (captioning). With this encoder-decoder architecture, the CNN and RNN can be trained end-to-end, inputting an image and outputting an ordered list of labels. Existing work differs slightly in how the CNN and RNN models are interfaced (see Figs.~\ref{fig:compare}(a)-(c)). However, they share a key characteristic: the image embedding that provides the CNN-RNN interface  is the final feature layer of the CNN  \cite{basic_model,mrnn,cnnrnn} (e.g. the FC7 layer of  Alexnet \cite{krizhevsky2012imagenet} or the final pooling layer of  GoogLeNet \cite{szegedy2015going}) or its linear transform \cite{showandtell,nic}.

Using such layers as the input to the RNN  has a number of adverse effects on learning an end-to-end recurrent image annotation model. First, since the CNN output feature is not explicitly semantically meaningful, both the label prediction and label correlation/grammar modelling tasks now need to be shouldered by the RNN model alone. This exacerbates the already challenging task of RNN training, since the number of visual concepts/words is often vast (there are more than 12,000 words in the MS COCO training captions) and their correlation is rich.  Second, a connected CNN-RNN model is effectively rather deep considering the RNN unrolling; existing CNN-RNN models apply supervision only at the final RNN output and propagate the supervision back to the earlier (CNN) layers. This leads to training difficulties in the form of ``vanishing'' gradients \cite{lee2015deepSupNet}.  In addition, joint training of CNN and RNN has to be  carried out very carefully to prevent noisy gradients  back propagated from the RNN from corrupting the CNN model. As a result, model convergence is often extremely slow \cite{nic}.


In this paper we propose to change the image embedding layer and introduce semantic regularisation to a CNN-RNN model in order to produce significantly more accurate results and make model training more stable and faster. Specifically, we perform multi-task learning where the auxiliary task (besides tagging/sentence generation) is to regularise the image embedding/interface layer to encode semantically meaningful visual concepts which are directly related to the label prediction task (Fig.~\ref{fig:compare}(d)). This can be understood from several perspectives: (i) As splitting up the system into a model for generating unary potentials (the CNN) by predicting the label individually, and modelling their relations (RNN) for structured prediction.  With the unary CNN taking the responsibility of concept prediction, the relational RNN model is better able to focus on learning concept correlations/sentence generation. In the multi-label classification case, where the label space of the semantic regularisation and the RNN output space are the same, this can be seen as analogous to CRF decoding of a joint distribution \cite{zheng2015crf_rnn}. (ii) As a deeply supervised network \cite{lee2015deepSupNet}, providing auxiliary supervision to the middle of what is effectively a very deep network. Such deep supervision improves accuracy and convergence speed \cite{lee2015deepSupNet,szegedy2015going}. In our case specifically, it largely eliminates the problem of noisy RNN gradients back-propagating to corrupt the CNN encoder \cite{nic}. It thus allows for better and more efficient fine-tuning of the CNN module, as well as fast convergence in end-to-end training of the full CNN-RNN model.   (iii) As pursuing an encoder-decoder model with prior bias of preferring semantically meaningful codes \cite{Attribute2Image}.

The contributions of this paper are as follows:
(1) We propose a novel CNN-RNN image annotation model which differs from the existing models in the selection of the image embedding layer and in the introduction of deeply-supervised semantic regularisation to the embedding layer. (2) Our proposed semantic regularisation enables reliable fine-tuning of the CNN image encoder as well as the fast convergence of  end-to-end CNN-RNN training. (3) We demonstrate through extensive experiments that on both multi-label classification and image captioning, we achieve the state-of-the-art performance.

\section{Related work}\label{sec:related_work}

\noindent\textbf{Deep multi-label classification} \quad  
Many earlier studies \cite{feifei} treat the multi-label classification problem as multiple single label classification problems and ignore the rich correlations in the label space. In order to model  label correlation, a structured output model is required. Deng \etal~\cite{hex} propose a hierarchy and exclusion graph (HEX) to model the structure of labels; however, they only focus on single label classification. Deep structured learning is widely employed in object segmentation. For instance, Zheng \etal~\cite{zheng2015crf_rnn} present an end-to-end structured model that combines the CNN model with a CRF. It allows for fast inference and learning of a deep model with Gaussian edge potentials. This was extended by Chen \etal~\cite{structure} to a deep model which combines MRFs and CNN to model output correlations, and is applied to multi-label classification. 
Multi-label structure was also effectively modelled by Conditional Graph Lasso \cite{graphlasso}, but for shallow models.

These CNN-CRF/MRF models work well for image segmentation. However, for multi-label classification, the large label space,  seriously imbalanced label distribution, and the need for variable length prediction challenge the application of these models \cite{cnnrnn}. Recently, the CNN-RNN~\cite{basic_model,cnnrnn} pattern has been applied to multi-label classification to capture label correlations, as well as address label imbalance and variable length prediction. Since RNN requires sequential input, before training the unordered label set is converted to an ordered list, \eg, frequent first~\cite{cnnrnn} or rare first~\cite{basic_model}. Small classes can be promoted by using the rare first order. For structured prediction, 
it is more computationally efficient than CNN-CRF, as it only iterates until the required number of labels are output.
Furthermore, it is an end-to-end predictive model as it outputs labels directly, rather than prediction scores, thus eliminating tricky  prediction score thresholding heuristics. Our model is related to \cite{basic_model,cnnrnn} in that it  follows the CNN-RNN design pattern; however, it uses a semantically regularised image embedding layer as the interface layer rather than an unregularised CNN  feature layer.


Another line of work is to incorporate side information in multi-label classification, since side information could be complementary to the image data. The side information could be user tags or groups from  image metadata \cite{sinn,feifei}. Johnson \etal~\cite{feifei} uses a non-parametric approach to find image neighbours according to the metadata, and then aggregates visual information of the image and its neighbours with a deep network to improve classification. In \cite{sinn} tags, groups, and labels are modelled by different concept layers, which corresponds to different level of abstractions. Messages can be passed top-down and bottom-up by leveraging a bidirectional structured network. Side information can also be exploited in our model, but we show that even using less side information, \eg, tags only, our model can outperform those in \cite{sinn,feifei} significantly.

\noindent\textbf{Neural network based image captioning} \quad 
A number of recent captioning studies take a bottom-up approach, where words or phrases are first detected and then composed to sentence with a language model. Fang \etal~\cite{mil} propose a caption model that first detects keywords using a multiple instance learning, and then uses the keywords to generate sentences. A similar model is proposed in \cite{Wu_2016_CVPR} with the main difference being that LSTM is used as the language model. Compared with these model, our model is an end-to-end CNN-RNN model which jointly learns the image encoding and language decoding modules. 

CNN-RNN based image captioning models have become popular. Vinyals \etal~\cite{showandtell,nic} follow an encoder-decoder scheme, and feed image features as the initial input to the RNN decoder, so that sentences are generated according to the image. A similar approach is employed in \cite{feifeicap}. Our work is related to \cite{showandtell}, but we use semantic concepts to regularise the representation of the CNN-RNN interface layer, which leads to significantly improved performance  and much easier model training.  Recently, visual attention has been incorporated to improve captioning accuracy. Xu \etal~\cite{Xu2015show} propose a model capable of sequentially attending to discriminative regions to improve the caption generation.  You \etal~\cite{attention} propose  to combine visual attributes and image features. An attention mechanism is introduced to reweight attribute predictions and merged with both the input and output of the RNN.
 Image features are fed at the first step as an external guide. Such attention models could easily be integrated into our model to further improve  performance. 

\vspace{0.1cm}\noindent \textbf{Semantic regularisation in deep encoder-decoders} \quad The idea of introducing semantic regularisation to an encoder-decoder model has been exploited in the context of image synthesis. Yan \etal~\cite{Attribute2Image} extend the variational autoencoder \cite{journals/corr/KingmaW13} by introducing attribute induced semantic regularisation to the middle embedding layer. A similar model based on generative adversarial networks is also proposed \cite{reed2016generative}.  Despite the similar strategy to ours, the objective is very different: we use the encoder-decoder architecture to align the text and image modalities and  middle-layer supervision is employed to achieve more effective and efficient training of both the encoder and decoder.  

\section{Methodology}\label{sec:method}

\begin{figure*}[htbp]
\begin{center}
\includegraphics[width=130mm]{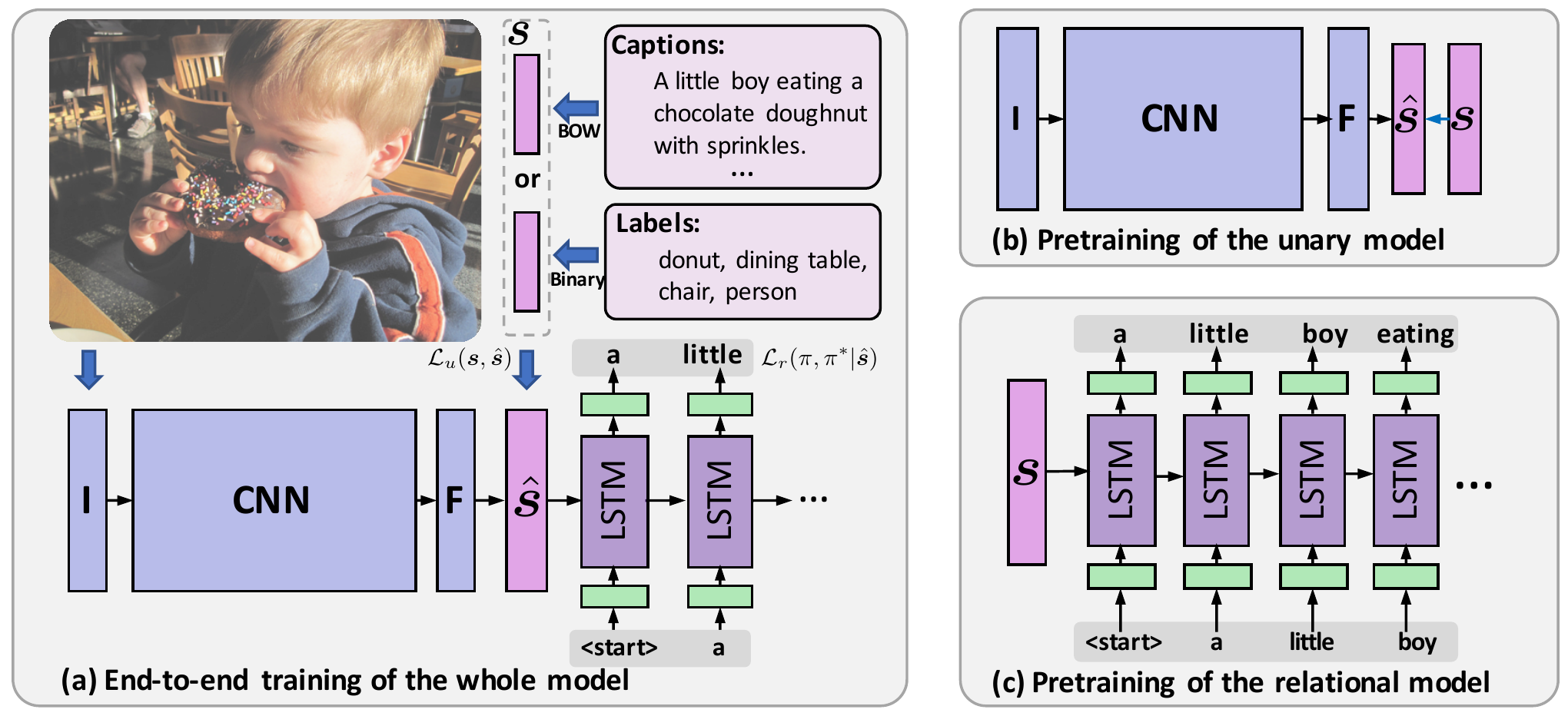} 
\caption{The full pipeline of the proposed semantically regularised annotation model. The ground truth semantic concepts  serve as strong supervision in the middle to regularise the training of the unary model (a). Due to the use of semantic concepts as the interface between CNN and RNN, the unary model and relational models can be pretrained in parallel, as shown in (b), (c).} \label{fig:main}
\end{center}
\end{figure*}

We first give an overview of existing CNN-RNN models before introducing our semantically regularised CNN-RNN. Its  application to multi-label classification and image captioning are detailed in Sec.~\ref{sec:multi_label} and Sec.~\ref{sec:caption} respectively.  

\subsection{CNN-RNN}
 A CNN-RNN model is composed of two parts: a visual encoder perceives the visual content of an image and encodes it to an image embedding; and a decoder takes the embedding as input and  generates sequences of labels (words).
  
Given an image $I$, a visual encoder will encode it to a fixed length  vector $\bm{I}_{e} \in \mathbb{R}^{d \times 1}$ called image embedding:
\begin{equation}
\bm{I}_{e} = f_{enc}(I),
\end{equation}
\noindent where $f_{enc}$ is the encoder, which could be a pretrained CNN optionally with some additional transformation layers. So $\bm{I}_{e}$ could either be a feature layer \cite{basic_model,mrnn,cnnrnn}, \eg, FC7 layer of VGG16 \cite{vgg16}, or its linear transform \cite{showandtell,nic}. In this paper, we enforce it to be a semantic representation to better interact with the RNN.

The RNN decoder will then take $\bm{I}_{e}$ as a condition, and generate a predictive path $\bm{\pi}=(a_1,a_2,...,a_{n_s})$, where for multi-label classification, $a_i$ is semantic label, and $n_s$ is the number of labels predicted for image $I$; while for image captioning $a_i$ is the word token, and $n_s$ is the length of the sentence. The path is an ordered sequence, so in multi-label classification, a priority of the labels has to be defined to convert labels to a sequence. We take a rare first order so as to give rare classes more importance during the prediction, therefore countering the label imbalance problem.

Many different CNNs have been considered for the encoder, but for the RNN decoder, the long short-term memory (LSTM) model \cite{Hochreiter:1997:LSTM} has been chosen by almost all existing models. This is because it controls message passing between times steps with gates in order to alleviate the vanishing/exploding gradient problem which plagued the training of prior RNN models. The model has two types of states: cell state $\bm{c}$ and hidden state $\bm{h}$. 
Following \cite{lstm_cell}, a forward pass at time $t$ with input $\bm{x}_t$ is computed as follows. 
\begin{equation} \label{eq:lstm}
\begin{split}
&\bm{i}_t = \sigma(\bm{W}_{i,h} \cdot \bm{h}_{t-1} + \bm{W}_{i,c} \cdot \bm{c}_{t-1} + \bm{W}_{i,x}\cdot \bm{x}_t  + \bm{b}_i)\\
&\bm{f}_t = \sigma(\bm{W}_{f,h} \cdot \bm{h}_{t-1}+ \bm{W}_{f,c} \cdot \bm{c}_{t-1} + \bm{W}_{f,x}\cdot \bm{x}_t  + \bm{b}_f)\\
&\bm{o}_t = \sigma(\bm{W}_{o,h} \cdot \bm{h}_{t-1}+ \bm{W}_{o,c} \cdot \bm{c}_{t-1} + \bm{W}_{o,x}\cdot \bm{x}_t  + \bm{b}_o)\\
&\bm{g}_t = \delta(\bm{W}_{g,h} \cdot \bm{h}_{t-1} + \bm{W}_{g,c} \cdot \bm{c}_{t-1}+ \bm{W}_{g,x}\cdot \bm{x}_t  + \bm{b}_g)\\
&\bm{c}_t = \bm{f}_t \odot \bm{c}_{t-1} + \bm{i}_t \odot \bm{g}_t\\
&\bm{h}_t = \bm{o}_t \odot \delta(\bm{c}_t)
\end{split}
\end{equation}
where $\bm{c}_t$ and $\bm{h}_t$ are the model's cell  and hidden states, $\bm{i}_t$,~$\bm{f}_t$,~$\bm{o}_t$ are the activation of input gate, forget gate, and output gate respectively; $\bm{W}_{\cdot,h}$, $\bm{W}_{\cdot,c}$ are the recurrent weights, and $\bm{W}_{\cdot,x}$ is the input weight, and $\bm{b}_{\cdot}$ are the biases. $\sigma(\cdot)$ is the sigmoid function, and $\delta$ is the output activation function.

At time step $t$, the model uses its last prediction $a_{t-1}$ as input, and computes a distribution over  possible outputs:
\begin{equation} \label{eq:step}
\begin{split}
\bm{x}_t & = \bm{E}\cdot \bm{a}_{t-1}, \\
\bm{h}_t & = LSTM(\bm{x}_t, \bm{h}_{t-1}, \bm{c}_{t-1}), \\
\bm{y}_t & = softmax(\bm{W} \cdot \bm{h}_t + \bm{b}),
\end{split}
\end{equation}
where $\bm{E}$ is the word embedding matrix, $\bm{h}_{t-1}$ is the hidden state of the recurrent units at $t-1$, $\bm{W}$, $\bm{b}$ are the weight and bias of the output layer, $\bm{a}_{t-1}$ is the one-hot coding of last prediction $a_{t-1}$, and  $LSTM(\cdot)$ is a forward step of the unit. The output $\bm{y}_t$ defines a distribution over  possible actions, from which the next action $a_{t+1}$ is sampled.

To generate image-conditioned sequences, the decoder has to take advantage of the image embedding $\bm{I}_{e}$, and existing models achieve this in multiple ways.
 Vinyals \etal\cite{showandtell} (Fig.~\ref{fig:compare}(a)) propose to feed $\bm{I}_{e}$ as step zero input to the LSTM model, that is, $(\bm{h}_0,\bm{c}_0)=LSTM(\bm{I}_{e},\bm{0},\bm{0})$, where $\bm{0}$ is a zero vector. In this case the weights of the word embedding are shared with image embedding, which is a questionable assumption, as the two embeddings have very different meanings and their dimensions have not been aligned.  Instead of treating $\bm{I}_{e}$ as an LSTM input, Wang \etal \cite{cnnrnn} and Mao \etal \cite{mrnn} combine word embedding and image features via output fusion (Fig.~\ref{fig:compare}(c)). In contrast, Jin \etal \cite{basic_model} use the image embedding to initialise the LSTM (Fig.~\ref{fig:compare}(b)) by setting hidden state $\bm{h}_0=\bm{W}_i\cdot\bm{I}_{e}+\bm{b_i}$, where $\bm{W}_i$, $\bm{b_i}$ are image input weights and biases.

Despite these differences, existing CNN-RNN models have a key common characteristic: The image embedding $\bm{I}_{e}$ that acts as the interface between the CNN and RNN models is taken to be a layer of weak and implicit semantics, \eg, CNN feature layer, or its transform. This means that the RNN has to simultaneously learn to predict semantic concepts from the provided features, as well as model the correlation of those concepts. Learning to predict the concepts is harder for the RNN because gradients are back propagated from relatively `far' supervision away (the RNN outputs at future time steps). Moreover fine-tuning the CNN becomes tricky because noisy gradients propagated from the RNN can easily degrade rather than improve performance \cite{nic}.


\subsection{Semantically regularised CNN-RNN}
To reduce the burden on the RNN, we propose a divide-and-conquer strategy to separate two tasks:  semantic concept learning and relational modelling. Specifically,  semantic concept learning is now performed by the unary CNN model which takes as input images (and associated side information if any), and produces a probabilistic estimate of the semantic concepts. Relational modelling is handled by the RNN model which takes in the concept probability estimates and models their correlations to generate label/word sequences. Concretely, instead of using a CNN feature layer as embedding $\bm{I}_{e}$, we use the CNN label prediction layer, \eg, concept prediction layer of an Inception net \cite{inception}. Since the chosen embedding is trained under direct supervision of  ground-truth labels/visual concepts, it has clear semantic meaning: Each unit corresponds to a semantic concept.



As shown in Fig.~\ref{fig:main}, in our Semantically regularised CNN-RNN (S-CNN-RNN), the CNN part takes an image $I$ as input, and predicts the likelihood of the semantic concepts $\hat{\bm{s}} \in \mathbb{R}^{k \times 1}$ where $k$ is the number of semantic concepts\footnote{$k$ is the size of label space in multi-label classification. For image captioning, $k$ is the number of visual concepts, which is typically smaller than the vocabulary size as not all words are visual. }. The RNN model takes $\hat{\bm{s}} $ as input, and generates sequences $\pi$. The key implication is that supervision can now  be added at \emph{both} the RNN output layer and the embedding layer  $\hat{\bm{s}}$. This results in two losses: a loss for concept prediction $\mathcal{L}_u(\bm{s},\hat{\bm{s}})$ and a loss for relational modelling $\mathcal{L}_{r}(\pi,\pi^*|\hat{\bm{s}})$. 
Formally, we have
\begin{eqnarray}
\mathcal{L}_{u}(\bm{s}, \hat{\bm{s}}) &=& \sum_{i} \ell_u(\bm{s}_i, \hat{\bm{s}}_i) \nonumber \\ 
\mathcal{L}_{r}(\pi,\pi^*|\hat{\bm{s}}) &=& \sum_i \ell_r(\pi_i,\pi_i^* | \hat{\bm{s}}_i) \nonumber \\ 
\mathcal{L} &=& \mathcal{L}_{u}(\bm{s}, \hat{\bm{s}}) + \mathcal{L}_{r}(\pi,\pi^*|\hat{\bm{s}}) \label{eq:semantic},
\end{eqnarray}
where $\bm{s}_i$ is the ground truth concept labels for the $i$-th training image and $\hat{\bm{s}}_i$ is the corresponding prediction; For the RNN loss $\mathcal{L}_{r}(\pi,\pi^*|\tilde{\bm{s}})$, $\pi_i^{*}$ is the ground truth path; $\pi_i$ is the predicted path, which is a sequence of word tokens or list of tags. The specific form of the losses will be discussed next.



\subsection{Training and inference} \label{sec:train}
The introduction of semantic regularisation in the middle of CNN-RNN  allows for more effective and efficient model training. It facilitates a two-staged training strategy illustrated in Fig.~\ref{fig:main}.  In the first stage,  we pretrain the CNN model and RNN model in parallel and in the second stage, they are fine-tuned together.

\vspace{0.1cm}\noindent\textbf{CNN}\quad For pretraining of the CNN model (Fig.~\ref{fig:main}(b)), the ground truth semantic concepts $\bm{s}_i$ are used as the learning target in a standard cross entropy loss for $k$ visual concepts:
\begin{equation}
\ell_u(\bm{s_i},\hat{\bm{s}}_i)=\sum_j^k s_{ij}\cdot \log(\hat{s}_{ij}) + (1-s_{ij})\cdot \log(1-\hat{s}_{ij}),
\end{equation}
\noindent\textbf{LSTM}\quad 
For the LSTM pretraining (Fig.~\ref{fig:main}(c)), the concept input $\hat{\bm{s}}_i$ is first connected to  a fully connected (FC) layer before being used to set the initial hidden state of the LSTM\footnote{This is to allow for the flexibility of using arbitrary LSTM unit size.}. The LSTM model learns to maximise the likelihood of generating the target sequences conditioned on the semantic input, and the loss $\mathcal{L}_{r}(\pi,\pi^*|\hat{\bm{s}})$ is simply the sum of the negative log likelihood over all time steps. By feeding $\bm{s}$ , rather than $\hat{\bm{s}}$ the LSTM can be pre-trained independently of the CNN.

\vspace{0.1cm}\noindent\textbf{Joint CNN-LSTM}\quad 
After the CNN and RNN models are  pretrained, the whole model can be jointly trained by simultaneously optimising the deeply supervised joint loss $\mathcal{L}$. 
For inference, we condition on the image by setting the initial state, then feed a start signal and recurrently sample model predictions of the previous step as input until an end signal is generated. For multi-label classification, we just greedily  take the maximum model output, whilst beam search with a width of three is employed for image captioning \cite{nic}.

\section{Application to Multi-label Classification} \label{sec:multi_label}

\subsection{Formulation}

To apply our S-CNN-RNN to multi-label classification, we first rank the training labels according to their frequency in the training set and generate a ordered label list with the rare labels first. We also explore the use of side information \cite{feifei,sinn}: exploiting the noisy user-provided tags available with each image. In this case the model in Fig.~\ref{fig:main} is slightly modified.  Specifically, we pretrain a multiple layer perception (MLP)  (single 256 neuron hidden layer and ReLU activation) to predict the true tags given the noisy metadata. Then we combine the image model with the pretrained tag model by summing their predictions as the final embedding  $\hat{\bm{s}})$, and train them together with a cross entropy loss  \cite{qa_baseline}.

\subsection{Datasets and settings}

\noindent \textbf{Datasets} \quad Two widely  used large-scale benchmark datasets are selected to evaluate our model. \textbf{NUS-WIDE} \cite{nuswide} dataset contains 269,648 images. Originally coming from Flickr, there are 5,018 unique user tags released along with the images. Of them, 81 tags are manually selected and refined as the ground truth \cite{nuswide}, covering different aspects including object classes, scenes, and attributes. The ground truth labels are highly imbalanced: the most frequent tag, {\em sky} appears 74,190 times while the rarest one {\em map} appears 60 times. In addition, the user-provided tags are extremely noisy and sparse --  8.73 noisy tags per image on average.  Following \cite{feifei,sinn}, we consider two settings: multi-label classification with only imagery data and with both images and noisy tags as side information. The most popular 1,000 noisy user tags are kept and  we remove the images without any ground-truth tags. As in many Flicker based studies, the numbers of images used by different works vary as they download the images at different times. For fair comparison, we use the same train/test split ratio as \cite{feifei,sinn}; as a result, 15,000 images are used for training and 59,347 for testing. \textbf{Microsoft COCO} \cite{coco} is popular for tasks such as object detection, segmentation and image captioning. Following \cite{cnnrnn}, we also use it for multi-label classification by treating the 80 object classes as labels. Since there are normally many types of objects in each image,  it is naturally a multi-label classification problem. Because the label space contains objects only and  some objects are rather small, it is perhaps more suitable than NUS-WIDE for evaluating a structured prediction model, as modelling label correlation becomes more important to detect visually similar and small objects. We also download the original user tags from Flickr via the provided URLs, and the most frequent 1,000 tags are used as side information. We keep the original train/validation split \cite{coco} for training and evaluation.

\noindent \textbf{Implementation details} \quad For fair comparisons with  previous work,  in our S-CNN-RNN model, we use the caffe reference net \cite{caffe} as our unary CNN subnet on the NUS-WIDE dataset \cite{nuswide}, and VGG16  on MS COCO. Both models are pretrained on the ILSVRC12 dataset \cite{ilsvrc}. For pretraining the CNN subnet, the learning rate is set to 1e-4 for NUS-WIDE and 1e-3 for MS COCO. For the RNN subnet, we use 512 LSTM cells and a 256 dimensional word embedding. The output vocabulary size is set to 82 for NUS-WIDE and 81 for MS COCO, including all labels and an END token. We use the \verb|BasicLSTMCell| in TensorFlow as LSTM cells and employ ReLU as activation function. The relational model is trained using a RMS Prop optimiser with a learning rate of 1e-4. Both the code and trained models will be made available at the first author's website.

\noindent \textbf{Evaluation metrics} \quad
As in \cite{basic_model,cnnrnn}, both per-class and per-image metrics including mean precision and mean recall are used.  For each class/image, the precision is defined as: $p(\hat{y}, y) = |y\cap \hat{y}| / |\hat{y}|$; and recall is defined as: $r(\hat{y}, y)=|y\cap \hat{y}| / |y|$, where $y$ and $\hat{y}$ are the set of ground truth labels and predicted labels, and $|\cdot|$ is the cardinality of a set. The overall precision (O-P)/recall (O-R) is computed by taking the average precision/recall over all samples, while the per class precision (C-P)/recall (C-R) is averaged over all classes. F1 score is also computed by computing the harmonic mean of precision and recall. 
As in existing CNN-RNN models \cite{basic_model,cnnrnn}, we let the model to decide its own prediction length~\cite{basic_model,cnnrnn}, whilst for other compared fixed-length predictive models ~\cite{sinn,feifei,cnnrnn}, we use the top 3 ranked predictions.

\subsection{Experimental results}

\noindent \textbf{Competitors} \quad We compare with the following models. In all compared models, the same CNN and RNN modules are used. 
\textbf{CNN+Logistic:} This model treats each label independently by fitting a logistic regression classifier for each label. The results are reported in \cite{sinn}.
\textbf{CNN+Softmax:} A CNN model that uses softmax as classifier, and the cross entropy between prediction and ground truth is used as the loss function. The results reported in \cite{gong2013deep} for NUS-WIDE and \cite{cnnrnn} for MS COCO are used.
\textbf{CNN+WARP:} Same CNN model as above, but uses a weighted approximate ranking loss function for training to promote the prec@K metric. We use the results reported in \cite{gong2013deep} for NUS-WIDE and \cite{cnnrnn} for MS COCO. 
\textbf{CNN-RNN:} A CNN-RNN model which uses output fusion (Fig.~\ref{fig:compare}(c)) to merge CNN output features and RNN outputs \cite{cnnrnn}. 
\textbf{RIA:} In this CNN-RNN model \cite{basic_model}, the CNN output features are used to set the LSTM hidden state (Fig.~\ref{fig:compare}(b)).  Note that only smaller datasets were used in \cite{basic_model} and no code is available; we thus use our own carefully trained implementation in the experiments. 
\textbf{TagNeighbour:} It uses a non-parametric approach to find image neighbours according to metadata, and then aggregates image features for classification. Tag neighbour with 5K tags gives the best performance \cite{feifei}. It uses more side information than ours and is also transductive requiring access to the whole test set at once.
\textbf{SINN:} It \cite{sinn} uses different concept layers of tags, groups, and labels to model the semantic correlation between concepts of different abstraction levels. A bidirectional RNN-like algorithm is adopted to integrate information for prediction. 1K noisy tags and 698 query words are used as side information, which is more than what our model uses.
\textbf{Variants of our model:} Our S-CNN-RNN  with and without the side information are called Ours and Ours+Tag1K respectively.  Since the results reported by SINN  \cite{sinn} and TagNeighbour \cite{feifei} were based on ImageNet-pretrained  CNN models, for direct comparison we train a variant of our model that fixes the weights of the CNN subnet without finetuning (Ours+Tag1K Fix).

\vspace{0.1cm}\noindent \textbf{Results on NUS-WIDE} \quad We  make the following observations from the results shown in Table~\ref{tab:nuswide}. (1) The proposed S-CNN-RNN  performs consistently better than all alternatives in terms of the F1 score, both with (Ours+Tag1K) and without side information (Ours). (2) Looking at the precision and recall metrics, our model is more impressive on precision than recall. This is expected because compared to the non-CNN-RNN based models that predict a fixed number  of 3 labels, a CNN-RNN model tends to makes less predictions for this dataset with on average 2.4 ground truth tags per image. (3) The gaps between Ours and CNN-RNN \cite{cnnrnn} and  RIA \cite{basic_model}  show clearly the importance of adding semantic regularisation to the CNN embedding layer. (4) Comparing Ours+Tag1K Fix with TagNeighboor \cite{feifei} and SINN \cite{sinn}, we can see that significant improvements are obtained  even with less side information. This is due to the ability of the RNN decoder in our CNN-RNN model to model high-order label correlations. (5) Our full model (Ours+Tag1K) further improves over  Ours+Tag1K Fix on both per class and per image metric. This shows the importance of having an end-to-end CNN-RNN that can be trained effectively with the introduced deeply supervised semantic regularisation.  Qualitative results can be found in the supplementary material.



\vspace{0.1cm}\noindent \textbf{Results on MS COCO} \quad Similar conclusions can be drawn from the results in Table \ref{tab:coco}. Comparing with the results on NUS-WIDE, it is noted that the performance gain obtained by using the 1K noisy tags as side information is smaller. This is because that the number of user-provided tags on COCO is smaller (2.93 vs.~6.10 per image with 1K unique tags).

\begin{table}[t]
\begin{center}
\footnotesize
\begin{tabular}{@{} l|ccc|ccc @{}}
\toprule
Algorithms      & C-R & C-P & C-F1   & O-R & O-P & O-F1\\ 
\midrule
CNN+logistic \cite{sinn}	& 45.03	& 45.60	& 45.31	& 70.77	& 51.32	& 59.50 \\
CNN+Softmax \cite{gong2013deep}	& 31.22	& 31.68	& 31.45	& 59.52	& 47.82	& 53.03 \\
CNN+WARP \cite{gong2013deep}	& 35.60	& 31.65	& 33.51	& 60.49	& 48.59	& 53.89 \\ 
CNN-RNN \cite{cnnrnn} & 30.40 & 40.50 & 34.70 & 61.70 & 49.90 & 55.20\\
RIA \cite{basic_model}  & 43.62	& 52.92	& 47.82	& 66.75	& 68.98	& 67.85 \\ 
TagNeighboor$^\dagger$ \cite{feifei}	& 57.30	& 54.74	& 55.99	& 75.10	& 53.46	& 62.46 \\ 
SINN$^\dagger$ \cite{sinn}	& 60.63	& 58.30	& 59.44	& \textbf{79.12}	& 57.05	& 66.30 \\  
\midrule 
Ours	& 50.17	& 55.65	& 52.77	& 71.35	& 70.57	& 70.96 \\  
Ours+Tag1K Fix$^\dagger$ & 58.52	& 63.51	& 60.91	& 77.33	& 76.21	& 76.77 \\
Ours+Tag1K$^\dagger$	& \textbf{61.73}	& \textbf{71.73}	& \textbf{66.36	}& 76.88	& \textbf{77.41}	& \textbf{77.15} \\ 
\bottomrule
\end{tabular}
\end{center}
\caption{Multi-label classification results on NUS-WIDE. Results that use side information are marked with superscript $\dagger$.}
\label{tab:nuswide}
\end{table}

\begin{table}[htbp]
\begin{center}
\footnotesize
\begin{tabular}{@{} l|ccc|ccc @{}}
\toprule
Algorithms      & C-R & C-P & C-F1   & O-R & O-P & O-F1\\ 
\midrule
CNN+logistic \cite{cnnrnn}	& 58.60	& 59.30 & 58.90	& 65.00	& 61.70 & 63.30\\ 
CNN+Softmax \cite{cnnrnn}	& 59.00	& 57.00	& 58.00	& 60.20	& 62.10	& 61.10 \\ 
CNN+WARP \cite{cnnrnn}	& 59.30	& 52.50	& 55.70	& 59.80	& 61.40	& 60.70 \\
CNN-RNN \cite{cnnrnn} & 55.60	& 66.00 & 60.40	& 66.40	& 69.20 & 67.80\\ 
RIA	\cite{basic_model} & 54.07	& 64.32	& 58.75	& 64.57	& 74.20	& 69.05 \\ 
\midrule
Ours  	& 59.83	& 67.40	 & 63.39 & 68.73	& 76.63 & 72.47\\
Ours+Tag1K$^\dagger$	& \textbf{63.13}	& \textbf{71.38}	& \textbf{67.00}	& \textbf{73.05}	& \textbf{77.41}	& \textbf{75.16} \\ 
\bottomrule
\end{tabular}
\end{center}
\caption{Multi-label classification results on Microsoft COCO.}
\label{tab:coco}
\end{table}

\begin{table*}[htbp]
\begin{center}
\footnotesize
\begin{tabular}{p{1.45cm} |p{0.7cm}<{\centering} p{0.7cm}<{\centering} | p{0.7cm}<{\centering} p{0.7cm}<{\centering} | p{0.7cm}<{\centering} p{0.7cm}<{\centering} | p{0.7cm}<{\centering} p{0.7cm}<{\centering} | p{0.7cm}<{\centering} p{0.7cm}<{\centering} | p{0.7cm}<{\centering} p{0.7cm}<{\centering} | p{0.7cm}<{\centering} p{0.7cm}<{\centering}}
\toprule
\multirow{2}{*}{Metric} 	& \multicolumn{2}{c}{B-1}	\vline &  \multicolumn{2}{c}{B-2} \vline 	& \multicolumn{2}{c}{B-3} \vline	& \multicolumn{2}{c}{B-4} \vline	& \multicolumn{2}{c}{METEOR} \vline	& \multicolumn{2}{c}{ROUGE} \vline & \multicolumn{2}{c}{CIDEr}   \\ 
 	& c5	& c40	& c5	& c40	& c5	& c40	& c5	& c40	& c5	& c40	& c5	& c40	& c5	& c40 \\ 
\toprule
MSRCap \cite{devlin2015language}	& 0.715$_{20}$	& 0.907$_{8}$	& 0.543$_{19}$	& 0.819$_{9}$	& 0.407$_{19}$	& 0.710$_{10}$	& 0.308$_{16}$	& 0.601$_{10}$	& 0.248$_{16}$	& 0.339$_{11}$	& 0.526$_{19}$	& 0.680$_{14}$	& 0.931$_{15}$	& 0.937$_{16}$ \\ 
mRNN \cite{mrnn}	& 0.716$_{18}$	& 0.890$_{20}$	& 0.545$_{18}$	& 0.798$_{20}$	& 0.404$_{20}$	& 0.687$_{20}$	& 0.299$_{21}$	& 0.575$_{20}$	& 0.242$_{26}$	& 0.325$_{25}$	& 0.521$_{23}$	& 0.666$_{24}$	& 0.917$_{18}$	& 0.935$_{17}$ \\ 
V2L	\cite{Wu_2016_CVPR} & 0.725$_{10}$	& 0.892$_{18}$	& 0.556$_{11}$	& 0.803$_{17}$	& 0.414$_{14}$	& 0.694$_{17}$	& 0.306$_{18}$	& 0.582$_{18}$	& 0.246$_{19}$	& 0.329$_{21}$	& 0.528$_{16}$	& 0.672$_{18}$	& 0.911$_{20}$	& 0.924$_{20}$ \\ 
NICv2 \cite{nic}	& 0.713$_{21}$	& 0.895$_{17}$	& 0.542$_{21}$	& 0.802$_{18}$	& 0.407$_{18}$	& 0.694$_{18}$	& 0.309$_{15}$	& 0.587$_{16}$	& 0.254$_{8}$	& 0.346$_{6}$	& 0.530$_{15}$	& 0.682$_{11}$	& 0.943$_{12}$	& 0.946$_{14}$ \\ 
ATT \cite{attention}	& 0.731$_{9}$	& 0.900$_{14}$	& 0.565$_{9}$	& 0.815$_{11}$	& 0.424$_{8}$	& 0.709$_{11}$	& 0.316$_{9}$	& 0.599$_{11}$	& 0.250$_{13}$	& 0.335$_{17}$	& 0.535$_{8}$	& 0.682$_{12}$	& 0.943$_{11}$	& 0.958$_{11}$ \\ 
\midrule
Ours	& \textbf{0.743}$_{5}$	& \textbf{0.917}$_{4}$	& \textbf{0.578}$_{5}$	& \textbf{0.840}$_{4}$	& \textbf{0.434}$_{6}$	& \textbf{0.735}$_{5}$	& \textbf{0.323}$_{6}$	& \textbf{0.621}$_{5}$	& \textbf{0.255}$_{7}$	& \textbf{0.343}$_{7}$	& \textbf{0.540}$_{6}$	& \textbf{0.691}$_{5}$	& \textbf{0.986}$_{6}$	& \textbf{1.002}$_{5}$ \\ 
\bottomrule
\end{tabular}
\end{center}
\caption{Results from the official MS-COCO testing server (\url{https://www.codalab.org/competitions/3221\#results}). The subscript indicates the ranking as on the submission date w.r.t. each metric.}
\label{tab:cap_test}
\end{table*}

\section{Application to Image Captioning} \label{sec:caption}	

\subsection{Datasets and settings}

\noindent \textbf{Datasets and metrics} \quad We use the popular Microsoft COCO dataset \cite{coco} for evaluation. The dataset contains 82,783 training images and 40,504 validation images. Each image is manually annotated with 5 captions. The comparison against the state-of-the-art is conducted using the actual MS COCO test set comprising 40,775 images. Note that the annotation of the test set is not publicly available, so the results are obtained from the COCO evaluation server. For an ablation study, we also follow the setting of \cite{showandtell,nic} by a held-out set of 4,051 images from the validation set as the test set. The widely used BLEU, CIDEr, METEOR, and ROUGE scores are employed to measure the quality of generated captions. For the ablation study, they are computed using the \verb|coco-evaluation| code \cite{DBLP:journals/corr/ChenFLVGDZ15}.  

\noindent \textbf{Implementation details} \quad For our S-CNN-RNN,  we use  Inception v3 \cite{inception} as the CNN subnet, and an LSTM network is used as RNN subnet. The number of LSTM cells is 512, equalling to the dimension of the word embedding. The output vocabulary  size for sentence generation is 12,000. Note that all these are exactly the same as the NIC v2 \cite{nic} model ensuring a fair comparison. For semantic regularisation by deep supervision of image embedding layer, we need to extract a set of semantic concepts/training labels from the vocubulary.  To this end, we follow \cite{mil} and simply use the 1,000 most frequent words in the captions, which cover 92\% of  word occurrences.  The ground truth labels for a training image is defined as the words that appear at least once in the 5 captions. For the CNN pretraining, we initially just learn the prediction layer, and then tune all the parameters for 30,000 iterations with a batch size of 32 and learning rate of 1e-4. In parallel, the RNN model is pretrained for 1,000,000 iterations with the ground truth semantic labels as image embedding. After both models are pretrained, the full model is fine-tuned for 500,000 iterations. 

\subsection{Experimental results}
\noindent \textbf{Competitors} \quad
Five state-of-the-art models are selected for comparison:
\textbf{MSRCap}: The Microsoft Captivator \cite{devlin2015language} combines the bottom-up based word generation model \cite{mil} with a gated recurrent neural network \cite{grnn} (GRNN) for image captioning.
\textbf{mRNN}: The multimodal recurrent neural network \cite{mrnn} uses a multimodal layer to combine the CNN and RNN.
\textbf{NICv2}: The NICv2 \cite{nic} is an improved version of the Neural Image Caption generator \cite{showandtell}. It uses a better image encoder Inception V3. In addition, scheduled sampling \cite{scheduled_sampling} and an ensemble of 15 models are used; both improved the accuracy of captioning. Neither is used in our model. 
\textbf{V2L}: The V2L model \cite{Wu_2016_CVPR} use a CNN based attribute detector to firstly generate 256 attributes, and then feed as initial input to a LSTM model to generate captions. 
\textbf{ATT}: The semantic attention model \cite{attention} uses both image features and visual attributes, and introduces an attention mechanism to reweight the attribute context to improve captioning accuracy. All five models use a CNN and a RNN, but only NICv2 does end-to-end training. In contrast, ATT does attention model and RNN joint training, and uses a 5-model ensemble. There is no joint training for the other three. 

\noindent \textbf{Results} \quad
We submit our results to the official evaluation server to compare with the five baselines which also appear in the official ranking. The evaluation is done with both 5 and 40 reference captions (C5 and C40).  It can be seen from  Table \ref{tab:cap_test} that our model beats all five competitors on all 14 metrics, often by a significant margin. Among the 39 submitted models, our model is ranked the 5th and we could not find references for the four higher ranked models. Note that our performance across all metrics is very consistent. In contrast, the 5 competitors often do well on some metrics but very badly on others. It is worth pointing out that  our result is obtained without a model ensemble, a practice commonly used in this type of benchmarking exercise (e.g., both NICv2 and ATT use ensembles). In addition, no auxiliary captioning data is used for training. This result thus represents the state-of-the-art. For qualitative results please see the supplementary material.

\noindent \textbf{Ablation study} \quad We compare our full model with two stripped-down versions. \textbf{NIC-F}: removing the semantic regularisation and use the CNN output feature layer as the inference $\bm{I}_e$   to RNN. This gives us the standard NIC model \cite{showandtell} with the same Inception v3 as CNN subnet. The model is finetuned end-to-end on COCO. 
\textbf{NIC-deeply}: this model is closer to ours -- it uses the same deeply supervised semantic regularisation as our model, but the penultimate feature layer is taken as the embedding, rather than the prediction layer $\hat{\bm{s}}$.
As a result, the CNN feature representation benefits from the deep supervision (rather than distal supervision via the RNN), but the specific embedding used as the RNN interface is not directly semantically meaningful.
The results on the validation set split are shown in Table \ref{tab:cap_dev}. It can be seen that: (1) Semantic regularisation is critical, e.g., it brings about 7\% on CIDEr comparing NIC-F and our full model. (2) The deep supervision is the most crucial contributor to the good performance of our model. \textcolor{black}{Even when the embedding layer is not semantically explicit as in NIC-deeply, the benefit is evident. The smaller gap between NIC-deeply and Ours is due to the use of the semantically explicit prediction layer as the embedding at the CNN-RNN interface.}


\begin{table}[htbp]
\begin{center}
\footnotesize
\begin{tabular}{@{} l|cccc @{}}
\toprule
Metric 	& CIDEr	& METEOR	& ROUGE	& B-4 \\ 
\midrule 
{NIC-F}	& 0.932	& 0.247	& 0.524	& 0.297 \\  
NIC-deeply & 1.006 &  0.258 & 0.543	& 0.323 \\
Ours & \textbf{1.054}	& \textbf{0.260}	& \textbf{0.550}	& \textbf{0.340} \\ 
\bottomrule
\end{tabular}
\end{center}
\caption{Ablation study results on the COCO validation set split. }
\label{tab:cap_dev}
\end{table}

\noindent \textbf{Computational cost} \quad
Thanks to the semantic regularisation, the proposed model can be trained very efficiently. The total training takes two days on a single Nvidia Titan X GPU. \textcolor{black}{In contrast training one of NIC's 15-model ensemble members takes more than 20 days on the same GPU.} In particular, the deep supervision allows the model to converge very fast.  For example, pretraining our Inception v3 \cite{inception} CNN  only needs 30,000 iterations with a batch size of 32. The pretraining of the RNN model is also fast since its inputs are ground truth labels. After the pretraining, the full model fine-tuning converges much faster than  NICv2. 
 

\section{Conclusion} \label{sec:conclusion}
We  proposed a semantically regularised CNN-RNN model for image annotation. The semantic regularisation makes the CNN-RNN interface semantically meaningful, distributes the label prediction and correlation tasks between the CNN and RNN models, and importantly the deep supervision makes training the full model more stable and efficient. Extensive evaluations on NUS-WIDE and MS-COCO demonstrate the  efficacy of the proposed model on both multi-label classification and image captioning.   


{\small
\bibliographystyle{ieee}
\bibliography{lstm}
}
\end{document}


\title{Supplementary Material for \\Semantic Regularisation for Recurrent Image Annotation\vspace{-1cm}}


\maketitle


\section{Qualitative results of multi-label classification}
\begin{figure}[hbt]
\begin{center}
\includegraphics[width=135mm]{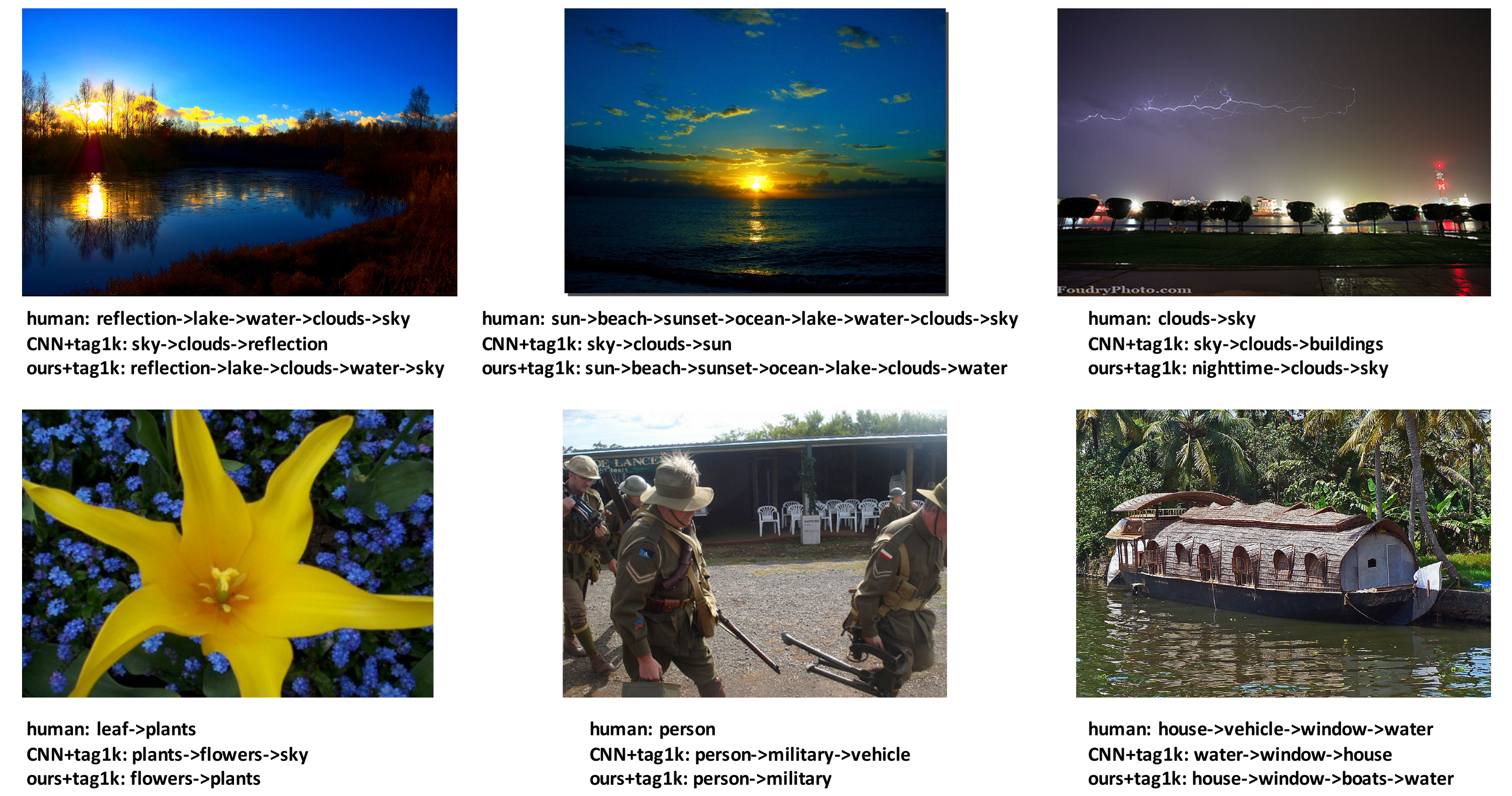} 
\includegraphics[width=135mm]{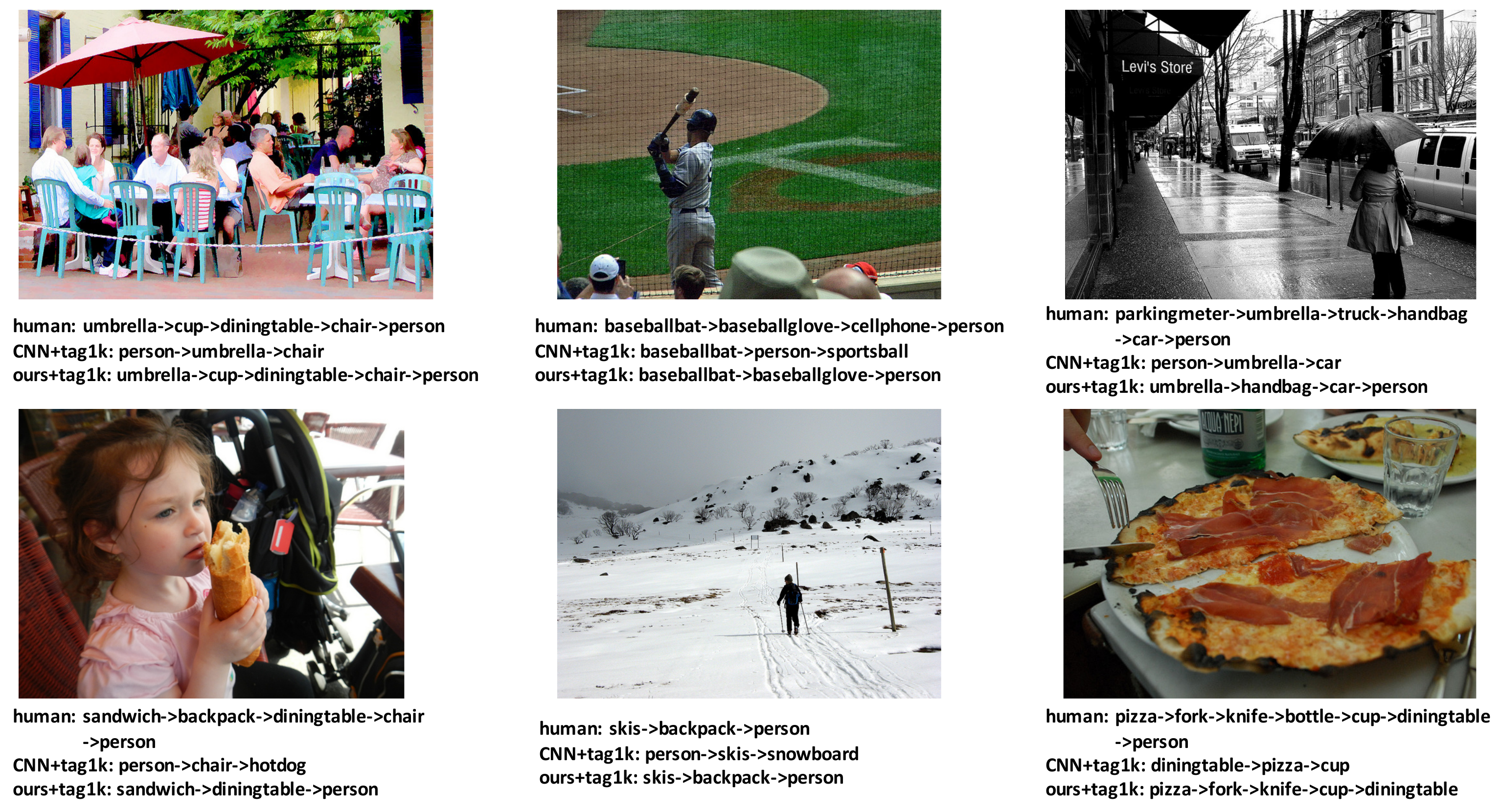} 
\caption{Qualitative results of multi-label classification. The top 6 images are from the NUS-WIDE dataset, and the bottom 6 are from MS COCO.} \label{fig:multi-label}
\end{center}
\end{figure}

Qualitative results of multi-label classification are shown in Fig. \ref{fig:multi-label}. The human row shows the ground-truth annotation, we organise them in a rare first order, where rare classes are presented earlier than the frequent classes. The CNN+tag1k use the model in \cite{qa_baseline}, the prediction are sorted according to their prediction scores in a descending order. The last row shows the results of our model, where the prediction order of RNN is preserved.

The results show that our algorithm mostly make predictions follow the desired rare-first order, thereby small classes are promoted. It tends to give more specific results rather than focus on large general concepts as does CNN+tag1k. Note that for some images, our prediction is even more accurate than ground truth, due to the missed tagging in manual labelling.

\section{Qualitative results of image captioning}
\begin{figure}[hbt]
\begin{center}
\includegraphics[width=135mm]{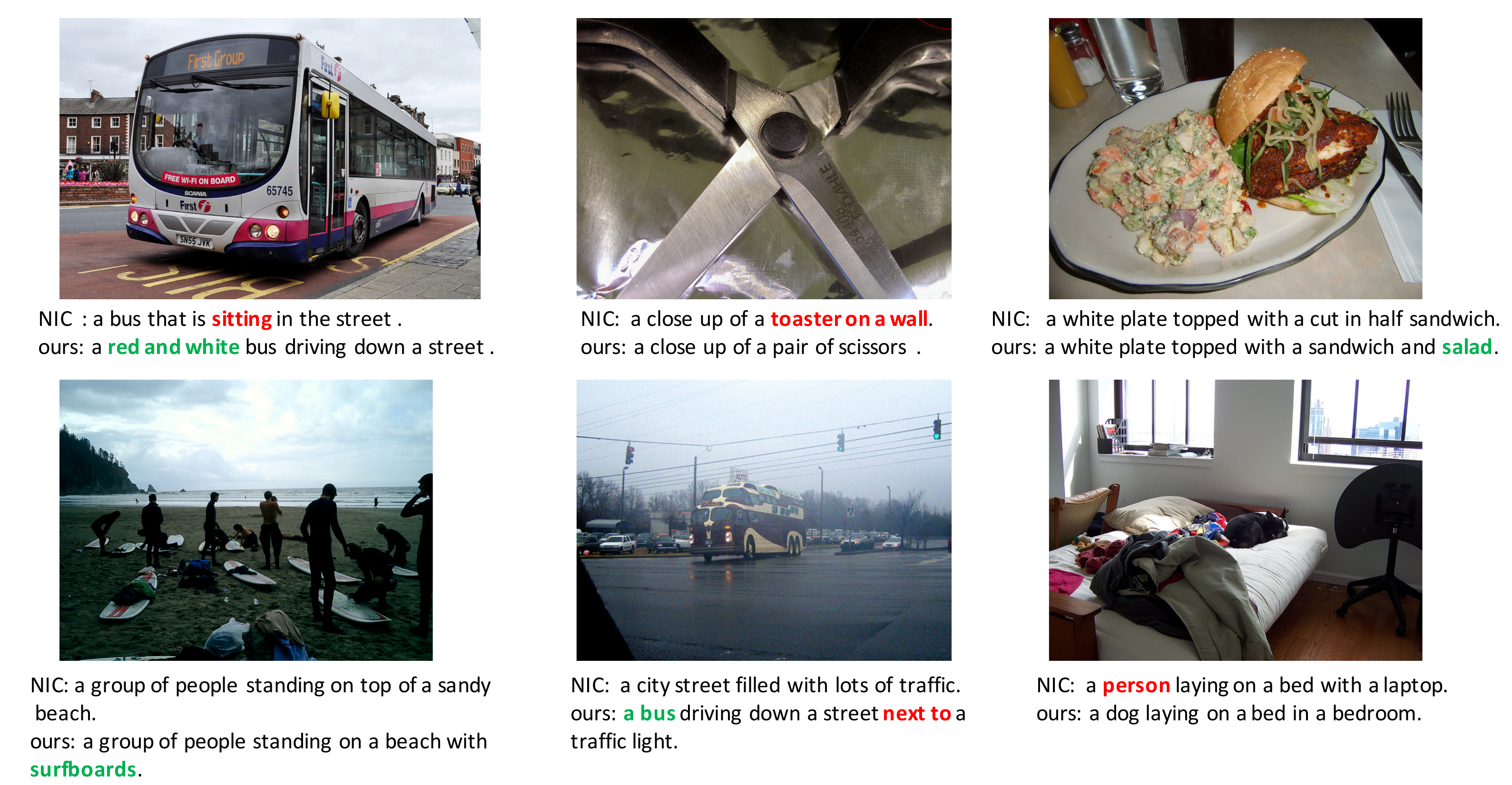} 
\includegraphics[width=135mm]{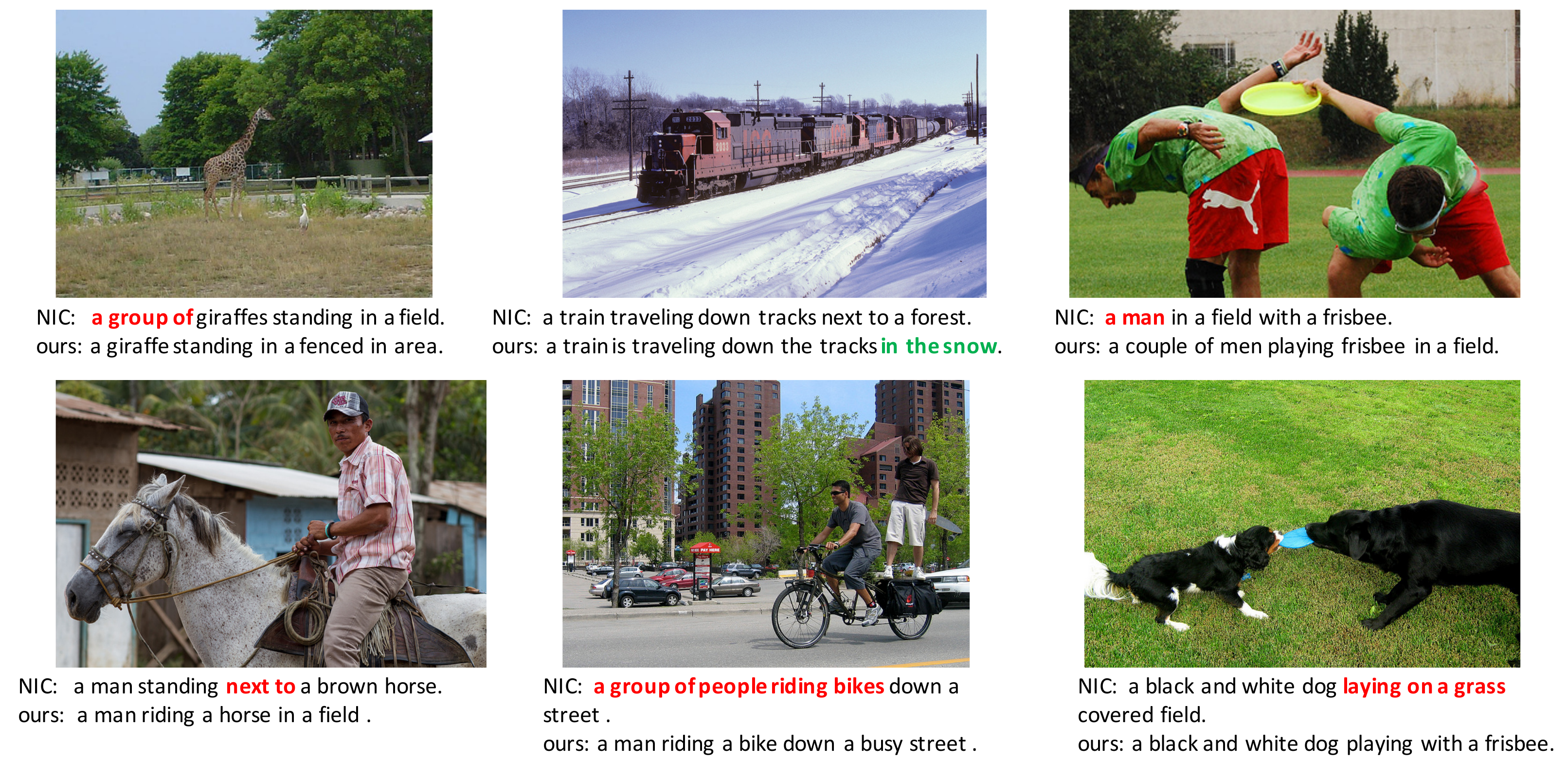} 
\caption{Qualitative results of image captioning on the MS COCO dataset. The errors in captions are hightlighted in red, while the fine-grained detials are hightlighted in green.} \label{fig:capgood}
\end{center}
\end{figure}

\begin{figure}[t!]
\begin{center}
\includegraphics[width=140mm]{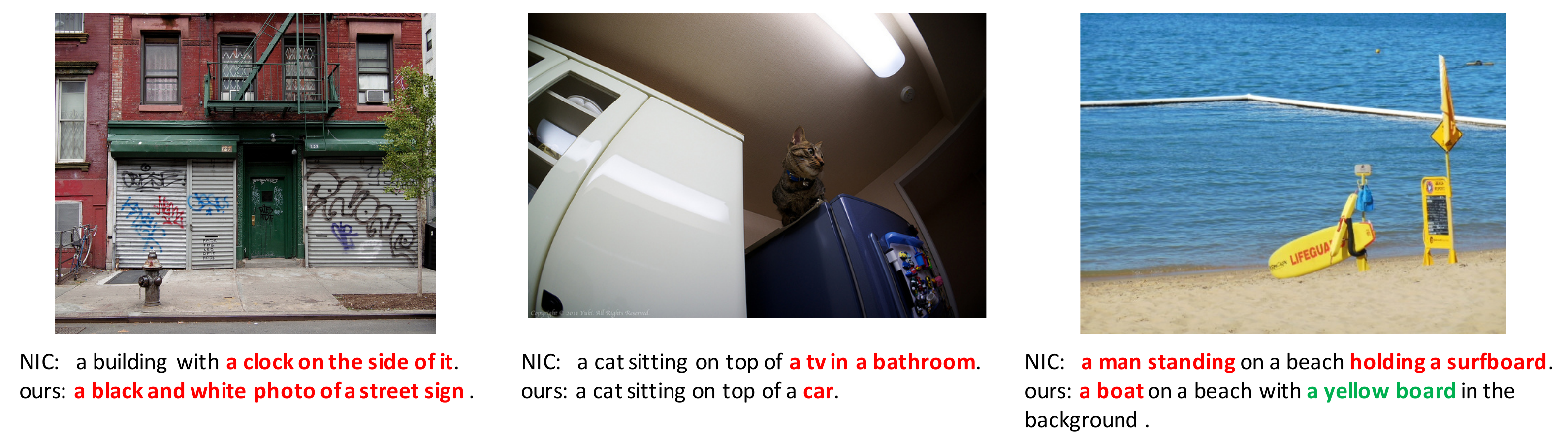} 
\caption{Failure cases of image captioning on the MS COCO dataset.} \label{fig:capbad}
\end{center}
\end{figure}

In this section, we show some example captions of our model and the NIC model \cite{nic}. The generated captions are shown in Fig \ref{fig:capgood}. Compared with the NIC model, our model is more accurate in recognising concepts, \eg, objects, colour, status, counts \etc, thus being able to capture object interactions and describe an image with more detailed nouns and adjectives. However, when the visual cue is compromised, our algorithm will also fail, as in the failure cases shown in Fig \ref{fig:capbad}. Novel concept can also influence captioning. The last example in Fig. \ref{fig:capbad} shows that novel object \emph{life guard station} is beyond the recognition ability of the algorithm, but it still manages to give a somewhat meaningful description.

{\small
\bibliographystyle{ieee}
\bibliography{lstm}
}